# Evaluation of Predictive Reliability to Foster Trust in Artificial Intelligence. A case study in Multiple Sclerosis.

Lorenzo Peracchio[1*], Giovanna Nicora[1*], Enea Parimbelli[1], Tommaso Mario Buonocore[1], Roberto Bergamaschi[2], Eleonora Tavazzi[2], Arianna Dagliati[1+], Riccardo Bellazzi[1+]

[1]Department of Electrical, Computer and Biomedical Engineering, University of Pavia, Italy
[2]IRCCS Mondino Foundation, 27100 Pavia, Italy

*Equal first authors.

+Equal last authors.

**Abstract**

**Background:** Applying Artificial Intelligence (AI) and Machine Learning (ML) in critical contexts, such as medicine, requires the implementation of safety measures to reduce risks of harm in case of prediction errors. Spotting ML failures is of paramount importance when ML predictions are used to drive clinical decisions. ML predictive *reliability* measures the degree of trust of a ML prediction on a new instance, thus allowing decision-makers to accept or reject it based on its reliability.

**Methods:** To assess reliability, we propose a method that implements two principles. First, our approach evaluates whether an instance to be classified is coming from the same distribution of the training set. To do this, we leverage Autoencoders' (AEs) ability to reconstruct the training set with low error. An instance is considered Out-of-Distribution (OOD) if the AE reconstructs it with a high error. Second, it is evaluated whether the ML classifier has good performances on samples similar to the newly classified instance by using a proxy model.

**Results:** We show that this approach is able to assess reliability both in a simulated scenario and on a model trained to predict disease progression of Multiple Sclerosis patients. We also developed a Python package, named *relAI*, to embed reliability measures into ML pipelines.

**Conclusion:** We propose a simple approach that can be used in the deployment phase of any ML model to suggest whether to trust predictions or not. Our method holds the promise to provide effective support to clinicians by spotting potential ML failures during deployment.

**Corresponding author**: giovanna.nicora@unipv.it

# 1. Introduction

Machine Learning (ML) techniques are witnessing an increase in their application in the most diverse settings; among all, healthcare is drawing considerable interest, in light of the large amount of clinical, genomic and sensor data that are regularly produced during clinical processes, and that promote the implementation of Artificial Intelligence (AI) tools [1]. Whilst a wide range of opportunities for improving the quality of patient care is provided by the clinical use of AI systems, it must be stressed that it unavoidably implies potential harm [2]: erroneous outputs, such as those of Clinical Decision Support systems [3], may lead to severe damage to patients, causing a lack of application of AI in clinical practice despite the thousands of papers that are published each year. To overcome clinician skepticism, AI tools need to make well informed decisions for each patient, whose individual distinctiveness and differences are undervalued by models anchored on population-level accuracy metrics [4]. This highlights the need for trustworthy AI systems, urgency that was also stated in 2020 by the European Union with the publication of the Assessment List for Trustworthy AI (ALTAI) [5], a list incorporating ethical guidelines and requirements that need to be fulfilled to achieve the trustworthiness of AI systems. In this context, the term "trustworthy" is related to a set of seven social-technical properties that are translated by ALTAI into a checklist for self-assessment, which guides developers, deployers and users to benefit from AI without being exposed to unnecessary risks. Other sets of well-known principles promoting the use of an innovative and trustworthy AI have been outlined in 2019 by the Organization for Economic Co-operation and Development (OECD) [6], and in 2021 by the FUTURE-AI consortium, which defined guiding principles and best practices for trustworthy AI in medical imaging through an iterative process comprising an in-depth literature review, a modified Delphi survey, and online consensus meetings [7]. Each of the aforementioned lists, among the others, stress the principles of fairness, explainability, and robustness as determining to move towards a trustworthy use of AI systems. An overview of the techniques, issues, and challenges related to the application of the properties of trustworthy AI in healthcare can be found in [8]. Within the technical robustness and safety requirement of ALTAI, the concept of Reliability has been emphasized as crucial: following the work of Saria et al. [9], we use the term Reliability to indicate the degree of trust of the prediction of a ML model on a single unseen instance. Typically, a ML model is evaluated on the basis of performance metrics computed on test set samples: in a general classification problem, the most widespread metrics (i.e. Accuracy, Area Under the Curve (AUC), Brier Score, etc.) depend on the number of correctly and incorrectly classified samples, giving a portrait of the model's prediction capability. Similar performances are also expected in the deployment phase, when the model is routinely used to predict the class of new samples; yet, in reality, this is not always true. In particular, ML models may suffer from dataset shifts, i.e. changes in distribution between the training set and the new samples to be classified, caused by non-stationary environments resulting from a change of a spatial or temporal nature [10]. Authors in [8] highlight data distribution shifts as a major challenge that hinders the practical application of AI systems in real clinical settings; biomedical and health data can be subject to significant variations in the real world, which can affect the performance of AI tools. This situation arises, for instance, when a ML model is trained on clinical data collected at a specific time in a certain hospital and used to predict the class of samples collected considerably later, when the data collection protocol might have changed, or coming from different hospitals, potentially attended by patients with diverse socio-demographic characteristics. In other words, multiple sources of variation may impact the AI tool's robustness, such as data-, equipment-, clinician-, patient- and centre-related variations [7]. ML model's performance on new data that do not come from the same distribution as the training set could drastically fail, or its predictions may not be trusted [11]. As a result, falsely optimistic estimates of model performance that are unachievable in real-world clinical settings is a contributing factor to the non-use of predictive models in deployment [12]. Moreover, dataset shifts may strongly contribute to fairness violation, as most of the works that improve the algorithmic fairness are under the assumption of an identical training and test distribution [13,14]. In addition, while the metrics used during testing are useful tools for the general assessment of the model's performance, in deployment we may be interested in understanding how the classifier performs on individual cases, for which we usually do not have an available ground truth class: reporting the reliability of individual-level predictions is crucial when a model is applied in practice [15]. Hence, not only is it essential to design AI systems to be robust against real-world shifts, but also to implement approaches for the evaluation of their predictive reliability when deployed and applied on new unseen samples, as well as strategies for mitigation [16].

The reliability of AI systems is often approached from the standpoint of uncertainty, a concept which is typically perceived as a statement about the standard probability of a single prediction [17]. Authors in [17] highlight the need for a trustworthy representation of uncertainty as a key feature in ML, especially in critical domains such as healthcare.

Various methodologies have been developed to quantify the predictive uncertainty [18]. In general, uncertainty quantification methods evaluate the uncertainty level of a prediction to determine its reliability: the higher the uncertainty, the more unreliable the prediction is. Several of the proposed approaches quantify uncertainty either through confidence intervals around a prediction or by computing a single score. Uncertainty can be exploited to assess the reliability of a classifier, assuming to trust the "certain" predictions of a model.

Current approaches to assess Reliability are based on the posterior predicted probability or are classifier-related [19]. However, this can bring misleading and biased evaluations since a classifier's trustworthiness should not be directly judged by the classifier itself [20]: a high classification probability is not necessarily related to a high probability of a correct classification, for example in case of bad calibration of the classifier or in presence of dataset shifts. In [19] we presented an approach for the evaluation of pointwise (i.e., individual) reliability. This method relies on the principles introduced by Saria et al. in [9], i.e. the Density and the Local Fit Principles. When the class of a new sample is predicted by a classifier, the Density Principle checks whether such sample is close to the training distribution. This problem is typically addressed by Out-of-Distribution (OOD) detection methods, which aim at identifying data not generated by the same distribution of the training set, whose predictions may not be considered reliable [21]. On the other hand, the Local Fit Principle checks whether the classifier was accurate on the training samples closest to the new instance. Our previous method implements the Density Principle by retrieving the training samples and by comparing the new instance with the so-called "training border" to understand if the sample is OOD (see [19]). To implement the Local Fit Principle, our previous approach computes the accuracy of the classifier on a set of training samples similar to the new instance: if the accuracy is high, the predictions are more likely to be reliable. This method can be used to estimate reliability of every type of supervised models. Yet, it shows a major limitation: it assumes that the training data are available to the user, even during the deployment phase, when the model can be potentially used by different institutions. This brings privacy issues that make this method not always feasible, as privacy preservation is an open problem in ML [22], especially in healthcare [23]. Even if we encrypt the training data to preserve privacy, data owners may not want to share their data with external users. Additionally, it may be impractical to share the training set when its dimension is large.

Overparametrized Autoencoders (AEs) have shown to implement associative memories, storing training samples as attractors [24]: in a dynamical system, attractors represent physical properties towards which a system tends to evolve, and therefore overparameterized AEs can be seen as a promising technique to encapsulate the training set distribution without actually keeping memory of the training set. Recently, denoising diffusion probabilistic models (DDPM) have been exploited for OOD detection; to classify OOD inputs, authors in [25] proposed to noise them to a range of noise levels, reconstruct them through denoising autoencoders, and evaluate the resulting multi-dimensional reconstruction errors. Similarly, in [26], a diffusion probabilistic model learned over the in-domain data is employed. Given an image to be tested, the image is lifted off its original manifold through corruption and mapped to the in-distribution manifold with the trained diffusion model; if the image is mapped to a different manifold, or in general if a large distance emerges between the original image and the mapped image, then it is considered out-of-distribution. The main limitation of these methods can be found in their high computational demand, which makes it onerous to apply them in real-time OOD detection and thus limits their adoption.

We propose a new method to assess ML reliability with the aim to overcome the limits of the existing approaches; we want our method (1) to be independent from the model chosen for classification, (2) not to rely on the predicted posterior probability or on intrinsic properties of the classifier, and (3) to be able to evaluate the individual reliability without any need of keeping memory of the training data. Moreover, our main goal is to encourage the adoption of reliability assessment methods in a typical ML framework: the final aim will be providing health care professionals not only with a ML classifier but also with a tool able to assess the reliability of the predictions at run-time, which is crucial to boost their confidence in the application of ML systems. In the following, we introduce our novel method for the assessment of the individual reliability, developed and validated both on simulated and clinical data, based on Autoencoders [27]. Furthermore, we present *relAI*, a Python package complete with a set of functions making it easier for users to implement our method, hence to evaluate ML reliability. The workflow of our reliability assessment method is shown in Figure 1.

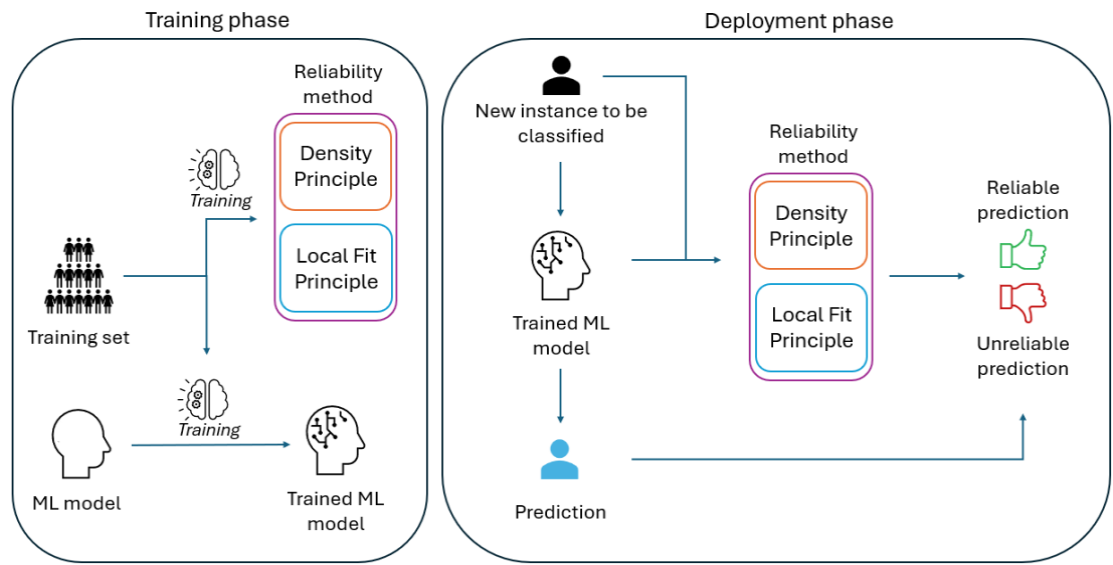

*Figure 1.* Reliability workflow.

## 2.Materials and Methods

In this section, we present a novel method for the assessment of individual reliability of ML predictions, which is based on the Density and the Local Fit Principles [9]; this framework is evaluated on a simulated dataset and on a Multiple Sclerosis (MS) real dataset.

### 2.1 Density Principle

To implement the Density Principle we propose a method that relies on autoencoders [27]. The autoencoder architecture is built on two neural networks (NN): the first one, the "encoder", consists of a feed-forward NN that encodes the input in a latent space. The second one, the "decoder", is a feed-forward NN that reconstructs the input from the latent representation. Autoencoders are exploited to learn an informative representation of the input data through an unsupervised framework, and to reconstruct the given input as similarly as possible. Autoencoders are trained by minimizing the reconstruction error between the output (i.e., the last decoder layer) and the input (i.e., the first encoder layer). The reconstruction error can be estimated through the Mean Squared Error (MSE), defined as the average squared difference between the input and its reconstruction:

$$MSE = \frac{1}{n}\sum_{i=1}^{n} (Y_i - \hat{Y}_i)^2 \qquad (eq.1)$$

where $Y$ is the $n$-dimensional vector in input and $\hat{Y}$ is the reconstructed one. When applied to unseen data, since the autoencoder is trained to reconstruct the training distribution, the MSE will be low for instances close to the training distribution and high for instances far from it. Thus, the evaluation of the MSE between a given input and its autoencoder reconstruction allows to establish whether the input is close or not to the training distribution, which is exactly what the Density Principle asks for. To understand whether the autoencoder projection is similar to the input, a MSE threshold needs to be defined: instances whose MSE is equal or lower than the fixed threshold are considered close to the training distribution, hence their classifier's prediction can be reliable, as the classifier is exploited on data similar to those used for its training. Instances whose MSE is higher than the fixed threshold are considered distant from the training distribution, hence their classifier's prediction might be unreliable as the classifier is working outside its training distribution. Selecting an appropriate MSE threshold is crucial, as it regulates the discrimination between a reliable and an unreliable prediction: a low MSE threshold results in a strict definition of what has to be considered reliable, while a high MSE threshold leads to a more relaxed definition. See Algorithm 1 pseudo-code for a complete overview of our approach.

**Algorithm 1**: Density principle implementation

```
 1: Train a classifier clf(·) on the training set
 2: Train an autoencoder model AE(·) on the training set
 3: Define a MSE threshold T
 4: for any new input sample x do
 5:     ŷ_x = clf(x) {classifier's prediction}
 6:     x̂  = AE(x) {autoencoder's reconstruction}
 7:     MSE_x = (1/n) Σ_{i=1}^{n} (x_i − x̂_i)²
 8:     if MSE_x ≤ T then
 9:            ŷ_x → reliable (according to the density principle)
10:     else
11:            ŷ_x → unreliable (according to the density principle)
12:     end if
13: end for
```

## 2.2 Local Fit Principle

The second principle that we implemented, to understand whether the classification is reliable, is the Local Fit Principle. Even if the classifier is used on samples coming from the same training distribution (i.e. the Density Principle is verified), the classifier may be fitted poorly in that area of the features space. We introduce an approach able to implement the Local Fit Principle without delivering the training set to the final user; the general idea consists in building a proxy model able to classify a new sample as reliable or unreliable according to the Local Fit Principle. In this section, we indicate as "*clf*" the classifier for which we want to assess the reliability of its predictions on new unseen samples, and as "*h*" the proxy model used to assess reliability. Our method is based on four steps:

1) *Generation of a dataset of synthetic points, built upon the training set*. For this purpose, one feasible strategy is to add random normal noise, with different values of variance, to the samples of the training set used to train *clf*. This allows to produce a large number of points with the same distribution of the training set, which can be exploited for the training of a proxy model *h* specifically developed to learn the local performance of the classifier.

2) *Definition of an accuracy threshold*. The accuracy threshold is used in the third step to label the synthetic points generated in the first step, on the basis of the local performance of *clf*.

3) *Labelling of the synthetic points*. Given a synthetic point, if the *clf* accuracy of its k closest training samples is equal or higher than the accuracy threshold, then it is labelled as "reliable"; otherwise it is labelled as "unreliable". The value of the accuracy threshold strongly influences this stage: selecting a higher or a lower value allows being more or less restrictive on the local reliability, respectively.

4) *Training of a classifier*. This has the ultimate goal of classifying the *clf* prediction on a new sample as reliable or unreliable according to the Local Fit Principle. To do so, a proxy classifier *h* is trained in a supervised fashion upon the synthetic points that were previously labelled based on the local performances of *clf*, so that it learns the decision rules for the discrimination of the "Local-fit reliability". Given a new instance to be tested, its "Local-fit reliability" is determined by the output of *h*; this supports the evaluation of reliability without requiring the training set in deployment, as the proxy classifier *h* has been trained to learn how accurate the original classifier *clf* is in the feature space. The proxy classifier *h* employed may be, for instance, a Decision Tree (DT) [28], a Multilayer Perceptron (MLP) [29], or in general any classifier able to characterize the feature space with great performances. The Local Fit Principle implementation is summarized in Algorithm 2.

**Algorithm 2:** *Local Fit principle implementation*

1: Train a classifier $clf(\cdot)$ on the training set
2: Generate the dataset of the synthetic points
3: Define an accuracy threshold $T$
4: Define the number of neighbors $k$
5: **for** each synthetic point $sp$ **do**

6:     retrieve $nn$ {its $k$ nearest training neighbors}
7:     retrieve $y_{nn}$ {set of the true classes of $nn$}
8:     $\hat{y}_{nn} = clf(nn)$ {classifier's predictions of $nn$}
9:     $acc_{nn} = accuracy\_score(y_{nn}, \hat{y}_{nn})$
10:     **if** $acc_{nn} \geq T$ **then**
11:         label $sp$ as reliable (according to the local fit principle) {1}
12:     **else**
13:         label $sp$ as unreliable (according to the local fit principle) {0}
14:     **end if**
15: **end for**

16: Train a classifier $h(\cdot)$ on the synthetic points

17: **for** any new input sample $x$ **do**
18:     $\hat{y}_x = clf(x)$ {classifier's prediction of $x$}
19:     $\widehat{rel}_{\hat{y}_x} = h(x)$ {Local Fit reliability prediction}
20:     **if** $\widehat{rel}_{\hat{y}_x} == 1$ **then**
21:         $\hat{y}_x \rightarrow$ reliable
22:     **else**
23:         $\hat{y}_x \rightarrow$ unreliable
24:     **end if**
25: **end for**

## 2.3 Dataset

### 2.3.1 Simulated Dataset

To test our approach, we generated a two-classes dataset of 6000 samples with two features to assess the effectiveness in recognising reliable and unreliable predictions; in particular, we simulate a binary classification problem under dataset shift. The training set (Fig 2.a) is made up of 800 samples and is used to train the classifier and to implement our reliability assessment method, whose MSE and accuracy thresholds are tuned on a validation set (Fig 2.b) composed of 300 samples. Both the training and the validation samples come from the same distribution and are equally distributed in the two classes. The test set (Fig 2.c) is used to evaluate our method: it is characterized not only by samples coming from the same distribution of the training and the validation sets, but also by a cluster of samples "Out-of-Distribution". OOD samples are employed to test the ability to detect unreliability in terms of Density Principle. Furthermore, we generated the training, validation, and test sets so that a region of overlap between the classes is introduced, where we expect that a machine learning classifier will have low values of accuracy; this region will be exploited to test the ability of our method to detect unreliability in terms of Local Fit Principle. With regards to the classification problem, we exploited a Random Forest with 100 trees.

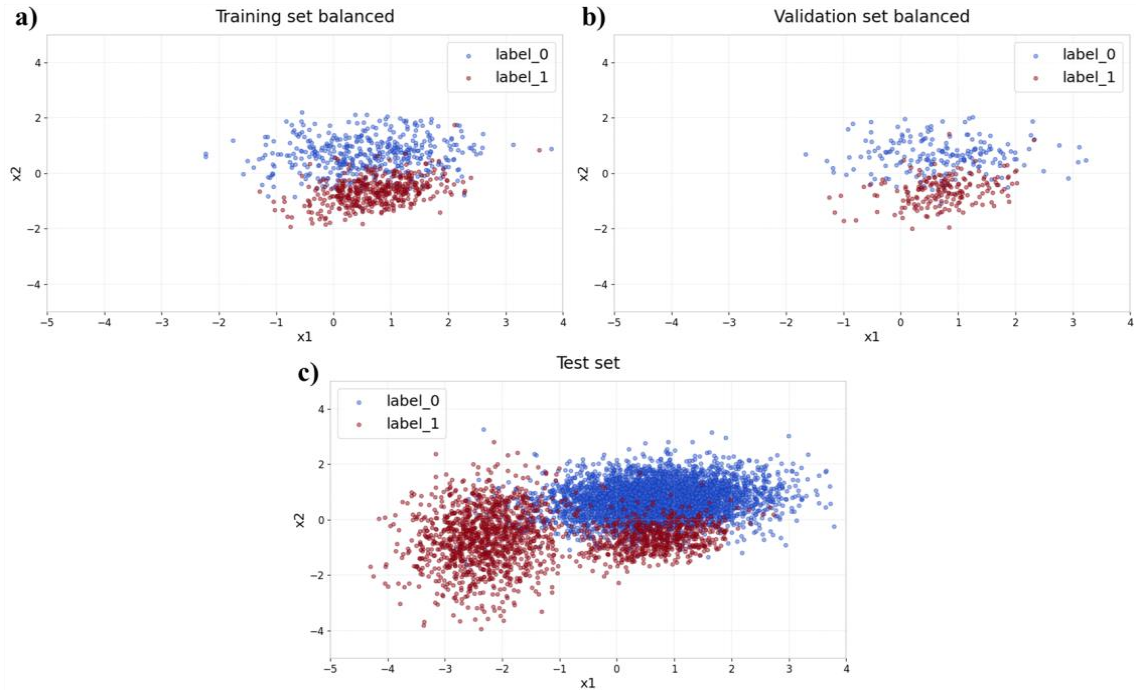

*Figure 2. a) Simulated 2D training set. b) Simulated 2D validation set. c) Simulated 2D test set.*

**2.3.2 Multiple Sclerosis Dataset**

To validate the method on a real clinical scenario, we applied it on a dataset of MS patients collected by the Mondino Hospital within the BRAINTEASER project, a "real world" dataset which offers two evaluation tasks focused on predicting the progression of Multiple Sclerosis. We focused on the task of predicting the disease's worsening, where worsening is considered if a patient crosses the threshold EDSS $\geq 3$ (Expanded Disability Status Scale [30]) at least twice within a one-year interval. From the training and test sets provided by the challenge, we excluded samples with missing data, and we only considered those collected from the clinical institution in Pavia, resulting in 392 training patients and 68 test patients. We only dealt with the static variables, which included demographics (sex, ethnicity, age, etc.), information about symptoms (brainstem symptoms, spinal cord symptoms, etc.), and information about the diagnostic delay and the time since the onset of the pathology. Moreover, following the data manipulation performed in [31], we extracted temporal EDSS trajectories as an additional feature by adopting the Latent Class Mixed Modelling (LCMM) approach. The preprocessing stage included factorization of the categorical features, normalization, and rebalancing of the minority class.

As previously mentioned, data distribution shifts are one of the main issues that can affect the reliability and performances of ML systems, preventing clinicians from actually applying them in clinical practice; thus, we introduce a temporal source of shift to the MS dataset, in order to test the ability of our proposed approach in identifying ML predictions on shifted samples as unreliable. To do so, we kept only the challenge's training patients whose diagnostic delay is equal or lower than 500 days (considered in-distribution), and we split them into a training set and a validation set used to train the classifier and to implement our method. With regards to the challenge's test set, 37 patients have an in-distribution diagnostic delay, while 165 of them are considered OOD, as their diagnostic delay is higher than 500 days. Table 1 summarizes the training, validation, and test sets' interquartile ranges of diagnostic delay. Further information on the experimental setups is available at https://github.com/bmi-labmedinfo/relAI, where we exploit the functions implemented in our python package relAI, published on the TestPyPi repository https://test.pypi.org/project/ReliabilityPackage/. (see relAI documentation at https://rel-doc.readthedocs.io/en/latest/#).

|  | # | Diagnostic delay (median [Q1-Q3]) |
|---|---|---|
| Training Set | 193 | 153 [70 - 304] |
| Validation Set | 65 | 122 [39 - 304] |
| Test Set | 202 | 1262.5 [554 - 2627] |

*Table 1. Number of patients in the training, validation, and test sets, and their interquartiles range of the diagnostic delay expressed in days*

As we processed the MS dataset to introduce a source of shift, the main goal of this section is to investigate the reliability assessment by the Density Principle. Since in the training set the feature of "diagnostic delay" is always lower than 500 days, fitting the Autoencoder also relying on this feature would result in a bad reconstruction for those test samples whose feature's value is significantly higher than 500, thus their "Density Principle" unreliability would be clearly detected. This approach is useful in deployment to detect features' values that have never been seen at the training stage, which may lead to distrust of the prediction. However, we want to go one step further by ignoring the "diagnostic delay" feature in the training of the classifier and the Autoencoder, to see if our method still detects the OOD test samples as "unreliable" just by learning the inherent relations between the other features' values when collected within 500 days, without directly relying on the variable of "diagnostic delay": we therefore excluded it in the training, validation, and test sets, together with the "time since onset" feature, as they are highly correlated. We designed the experiment by training a Random Forest with 500 estimators (classifier whose prediction reliability is under exam) on the in-distribution training set, upon which we also implemented our proposed reliability method. In particular, as we are not directly exploiting the value of diagnostic delay, we need to train the Autoencoder so that it accurately learns how to reproduce the training set: hence, we trained it for 10000 epochs and we selected the maximum reconstruction error of the training samples as the MSE threshold, so that all the training samples are considered in-distribution, hence reliable.

## 3. Results

## 3.1 Simulated Dataset

Fig. 3 shows the reliability of the Random forest's predictions on the test set, assessed by our proposed method. In this experiment, we fixed an Accuracy threshold of 0.85, in line with the Accuracy calculated on the Validation Set, and a MSE threshold set as the 98th percentile of the MSEs of the validation set; the number of neighbours used for the implementation of the Local Fit Principle was set to 5. Further details on the hyperparameters tuning can be found in Supplementary Materials, Appendix A. Accordingly to the Density Principle, 938/1200 (~78%) OOD samples are correctly detected as "unreliable", due to their distance from the distribution of samples used for training the classifier; those that are deemed as "reliable" lay in a region of the feature space where also some training points were distributed, hence they can be considered as not Out-of-distribution. In accordance with the Local Fit Principle, unreliability is also correctly identified in the region of overlap between the two classes, which is crossed by the decision boundary of the Random Forest, and therefore cannot be trusted due to the low performances of the classifier in that region. Ideally, as the reliable samples are more trustworthy, a classifier should perform better on this subset of data. Figure 4 shows the performance of the Random Forest on the entire test set, and on the reliable and unreliable subsets of the test set identified by our method; as revealed by the bar plots, the classifier's performance computed on the reliable and unreliable subsets are respectively better and worse than the ones computed on the complete test set.

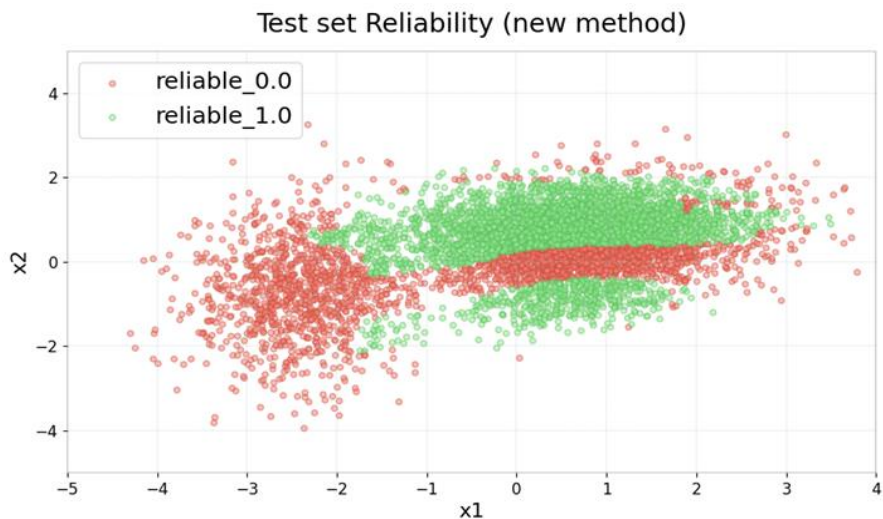

*Figure 3.* Reliability of the samples of the 2D simulated test set, assessed by the method proposed in this work; reliable samples in green, unreliable samples in red.

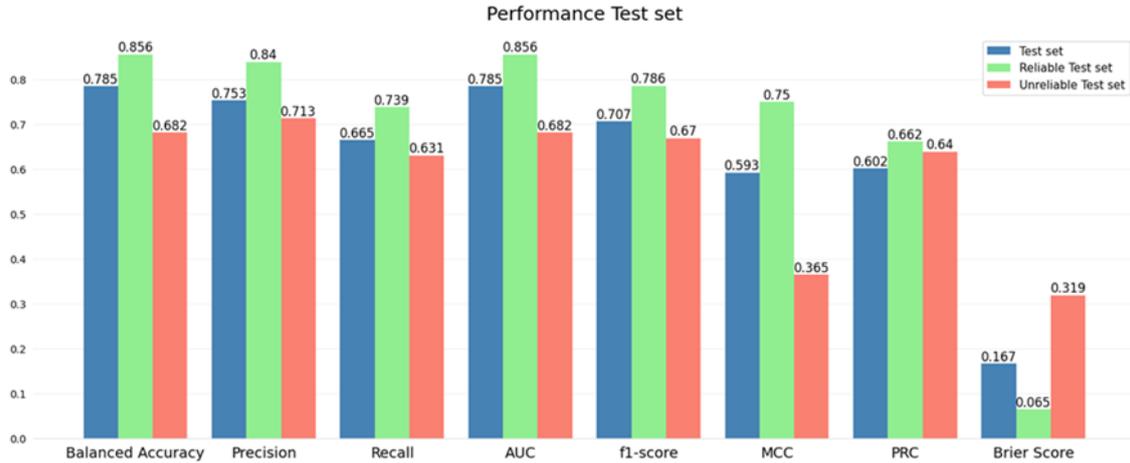

*Figure 4. Performance of the Random Forest on the whole Simulated 2D test set (in blue), and on its reliable (in green) and unreliable (in red) subsets identified by our method.*

To validate our method, we also performed a comparison with the uncertainty-based method proposed in [32] and with our previous implementation [19]; Fig.5a and Fig.5b show how the aforementioned approaches assess reliability on the test set.

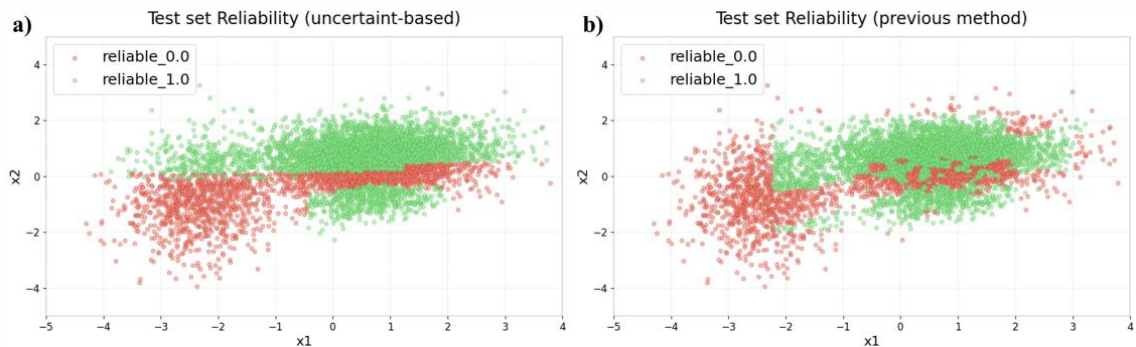

*Figure 5. Reliability of the samples of the 2D simulated test set, assessed by **a**) the uncertainty-based method proposed in [32] and **b**) our previous implementation [19]. Reliable samples in green, unreliable samples in red.*

As shown, both the uncertainty-based method and our previous approach are able to label as "unreliable" the samples that fall in the "low performance" region of the feature space (region of overlap between the classes, crossed by the decision boundary). With regards to the shifted samples, only those points whose $x_2$ value is roughly lower than 0 are assessed as unreliable by the uncertainty-based method, and only those whose $x_1$ value is roughly lower than -2 are assessed as unreliable by our previous approach, which results in an overly sharp subdivision of the OOD points in the reliable and unreliable subsets. Moreover, we compared the classifier's performances on the reliable and unreliable subsets identified by these methods and the ones proposed in this work, as reported in Table 2. The classifier's performances on the reliable and unreliable subsets identified by our new method are respectively better and worse than the ones computed on the ones identified by the other two methods, which makes our new method better at separating the reliable and unreliable performances, as highlighted by the deltas.

|  | Autoencoder-based reliability | | | Uncertainty-based reliability | | | Borders-based reliability | | |
|---|---|---|---|---|---|---|---|---|---|
|  | Reliable | Unreliable | Delta | Reliable | Unreliable | Delta | Reliable | Unreliable | Delta |
| Balanced Accuracy ↑ | 0.856 | 0.682 | <u>0.174</u> | 0.769 | 0.710 | 0.059 | 0.782 | 0.730 | 0.052 |
| Precision ↑ | 0.840 | 0.713 | <u>0.127</u> | 0.811 | 0.727 | 0.084 | 0.800 | 0.730 | 0.070 |
| Recall ↑ | 0.739 | 0.631 | 0.108 | 0.568 | 0.729 | -0.161 | 0.600 | 0.708 | -0.108 |
| AUC ↑ | 0.856 | 0.682 | 0.174 | 0.769 | 0.710 | 0.059 | 0.782 | 0.730 | 0.052 |
| F1 Score ↑ | 0.786 | 0.670 | 0.116 | 0.668 | 0.728 | -0.060 | 0.686 | 0.719 | -0.033 |
| MCC ↑ | 0.750 | 0.365 | 0.385 | 0.623 | 0.420 | 0.203 | 0.633 | 0.460 | 0.173 |
| PRC ↑ | 0.662 | 0.640 | 0.022 | 0.539 | 0.674 | -0.135 | 0.557 | 0.659 | -0.102 |
| Brier Score ↓ | 0.065 | 0.319 | <u>-0.254</u> | 0.103 | 0.289 | -0.186 | 0.105 | 0.270 | -0.165 |

*Table 2. Classifier's performance on the reliable (in green) and unreliable (in red) subsets identified by our new proposed approach based on autoencoders, by the uncertainty based method reported in* [32]*, and by our previous approach based on "borders-detection"* [19]*. Yellow columns represent the deltas of performance (reliable - unreliable), underlined if the biggest between the three methods.*

Altogether, our new reliability assessment method is more suitable for detecting the OOD samples, as well as better in obtaining higher performance deltas, improvements that are achieved without keeping memory of the training data and without relying on the classifier itself, thus overcoming the limitations of the currently existing approaches.

**3.2 MS Dataset**

The results of the application of the reliability assessment on the samples of the test set is shown in Table 3, where are reported the number of reliable and unreliable predictions, and their corresponding features' characteristics and p-values of tests performed to check if statistically significant differences exist between features' characteristics in the reliable and unreliable set. Notably, the median value of diagnostic delay for the "unreliable" samples is 1476 days (interquartile range 557.2 - 4033), while it is 879 days (interquartile range 563 - 1781.2) for the "reliable" samples, a statistically significant difference according to the Kruskal-Wallis test. Even if the feature was not directly included in the training and test sets, our method detects as "reliable" the samples whose diagnostic delay is lower, thus closer to the training set one: this makes the usage of autoencoders proposed in this work a powerful tool for checking if a new test instance is far from the training distribution, even when the variable responsible for the dataset shift is not explicit. Fig. 6 shows the boxplot of the diagnostic delay feature in the reliable and unreliable subsets identified by our approach, illustrating the significant difference between them. Other worthwhile considerations derived from Table 3 regard ethnicity and the residence classification: the predictions for all the non-caucasian patients, although only 5, were found

"unreliable", and the prediction for the town patients were found "reliable" at a significantly higher rate with respect to rural areas and city patients.

|  |  | Test set | Unreliable Test set | Reliable Test set | p-value |
|---|---|---|---|---|---|
| n |  | 202 | 140 | 62 |  |
| Sex , n (%) | Female | 141 (69.8) | 92 (65.7) | 49 (79.0) | 0.083 |
|  | Male | 61 (30.2) | 48 (34.3) | 13 (21.0) |  |
| Residence classification, n (%) | Towns | 108 (53.5) | 58 (41.4) | 50 (80.6) | <0.001 |
|  | Rural Area | 65 (32.2) | 56 (40.0) | 9 (14.5) |  |
|  | Cities | 29 (14.4) | 26 (18.6) | 3 (4.8) |  |
| Ethnicity, n (%) | Hispanic | 3 (1.5) | 3 (2.1) |  | 0.321 |
|  | Caucasian | 197 (97.5) | 135 (96.4) | 62 (100) |  |
|  | Black African | 2 (1.0) | 2 (1.4) |  |  |
| MS in pediatric age, n (%) | No | 188 (93.1) | 127 (90.7) | 61 (98.4) | 0.068 |
|  | Yes | 14 (6.9) | 13 (9.3) | 1 (1.6) |  |
| Age at Onset, median [Q1, Q3] |  | 27.5 [23, 33] | 25 [21.8, 32] | 30.5 [26, 34.8] | <0.001 |
| Spinal cord symptoms, n (%) | No | 166 (82.2) | 114 (81.4) | 52 (83.9) | 0.827 |
|  | Yes | 36 (17.8) | 26 (18.6) | 10 (16.1) |  |
| Brainstem symptoms, n (%) | No | 52 (25.7) | 36 (25.7) | 16 (25.8) | 1.000 |
|  | Yes | 150 (74.3) | 104 (74.3) | 46 (74.2) |  |
| Eye symptoms, n (%) | No | 130 (64.4) | 85 (60.7) | 45 (72.6) | 0.143 |
|  | Yes | 72 (35.6) | 55 (39.9) | 17 (27.4) |  |
| Supratentorial symptoms, n (%) | No | 141 (69.8) | 99 (70.7) | 42 (67.7) | 0.796 |
|  | Yes | 61 (30.2) | 41 (29.3) | 20 (32.3) |  |
| Other symptoms, n (%) | No | 189 (93.6) | 127 (90.7) | 62 (100) | 0.046 |

| | Epilepsy | 1 (0.5) | 1 (0.7) | | |
| | Sensory | 12 (5.9) | 12 (8.6) | | |
| Diagnostic delay, median [Q1, Q3] | | 1262.5 [554, 2627] | 1476 [557.2, 4033] | 870 [563, 1781.2] | 0.038 |

*Table 3. Summary characteristics of the test set and its reliable and unreliable subsets set as identified by our method; features for which there is a statistically significant difference between the reliable and unreliable subsets are underlined.*

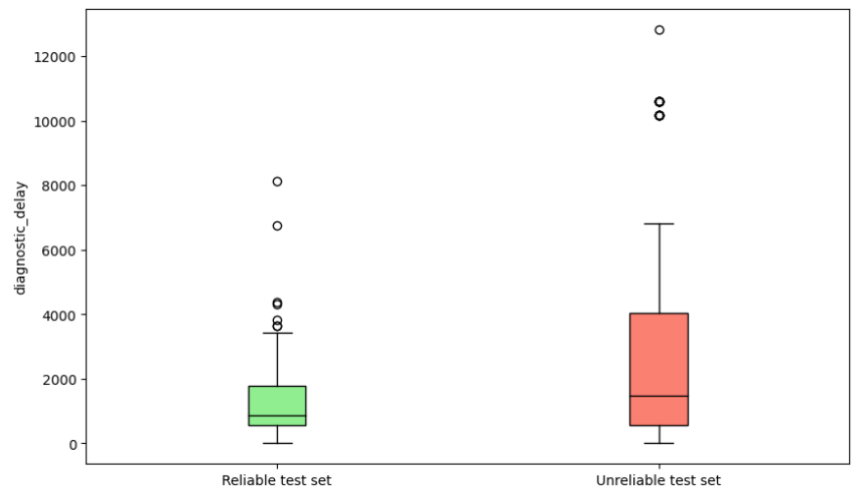

*Figure 6. Boxplot of the diagnostic delay feature for the reliable and unreliable subsets as identified by our method.*

The analysis of the classifier's performances on the reliable and unreliable subsets of the test set is finally executed in Figure 7: the performances on the entire test set (in blue) are poor, but if we only consider the reliable subset (in green) they show a considerable improvement, even if we are exclusively considering the Density Principle.

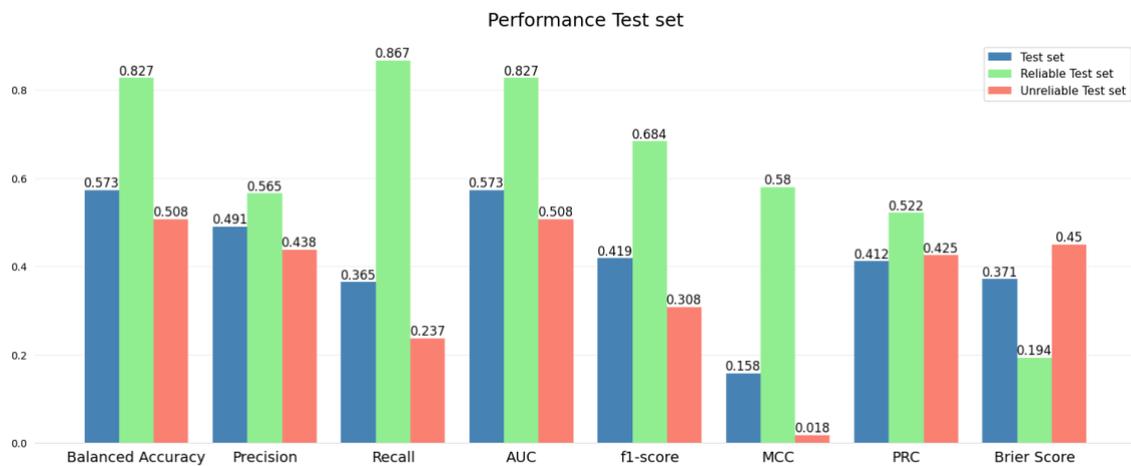

*Figure 7. Performance of the Random Forest on the MS test set (in blue), and on its reliable (in green) and unreliable (in red) subsets identified by our method.*

## 4. Discussion

Poor generalization of ML classifiers across different populations is raising concerns about their trustworthiness in high stakes applications, where their support to decisions can cause harm [11]. ML models are intended to be used on samples distributed as the training set, on which they usually perform well and can be seen as reliable systems. However, in deployment, ML models performances can decrease [33]. To identify potential failures at a single case resolution, different approaches can be exploited, such as the evaluation of ML uncertainty [34] or by using conformal prediction [35]. Contrarily to the aforementioned approaches, we propose a method to evaluate the pointwise reliability that is independent from the ML classifiers employed; in particular, we consider a prediction to be potentially unreliable when the predicted sample is not close to the training distribution (Density Principle reliability [9]), or when it lays in a low-accuracy region of the feature space (Local Fit Principle reliability [9]). Our approach relies on autoencoders to memorize training distributions, and on proxy models trained to learn low and high accuracy regions in the feature space. Tested on a simulated 2D dataset, our approach detected the unreliability of predictions both in terms of distance from the training distribution and inaccuracy of the classifier (Fig. 2). When applied in a real-life clinical scenario, it identified as unreliable the predictions on shifted patients even without directly relying on the feature causing the shift, proving the ability of this method to detect OOD samples with great performances (Fig. 5 and Table 3). Not only this allows us to detect the subset of samples for which the predictions are correct at a higher rate (Fig. 6), but such an ability can also be exploited to detect unfairness in ML systems. Fairness is one of the fundamental requirements expressed in the assessment lists for trustworthy AI; our method can be exploited for detecting unfairness, which represents an important obstacle to the dissemination of AI systems that can be considered trustworthy. However, some of the aspects of our approach need to be carefully considered: the implementation of the Density Principle requires the development of an autoencoder, whose architecture (number of layers, number of nodes for each layer, activation functions etc.) and training procedure need to be determined based on the specific application. However, our Python package *relAI* comes with a native function for the generation and training of an Autoencoder with default number of layers and nodes (depending on the input dimension), default activation functions and default training optimization parameters that produce an Autoencoder able to efficiently reproduce the training set. Moreover, our method relies on parameters that need to be tuned by the user: to assess the reliability in terms of Density Principle, a MSE threshold must be defined, while the number k of neighbours and the accuracy threshold are necessary for the implementation of the Local Fit Principle. The selection of these parameters reflects what is considered reliable and what is considered unreliable, thus may strongly influence the results. For further evaluation, more experiments need to be carried out, both on purpose-built simulated dataset and on real datasets. With the development of the *relAI* Python package we support the adoption of our method in real-world applications.

## 5. Conclusion

We implemented a method to assess ML prediction's reliability of single cases, used in deployment to identify potential failures and suggest to users whether or not to trust the predictions, and we made available our approach in a Python package (https://test.pypi.org/project/ReliabilityPackage/.) to promote the usage of safeguard measures in high-stakes applications, such as medicine. Future direction will include the formalization of our approach as a fairness tool, as well as the integration with explainability (XAI) methods, in order to provide an explanation in case of detection of an unreliable prediction, which would represent a significant improvement with respect to the simple rejection and a step ahead towards a trustworthy use of AI.

## Competing Interests

The authors have no conflicts of interests to declare

## CRediT author statement

Lorenzo Peracchio: Conceptualization, Methodology, Software, Writing-Original Draft, Visualization, Validation; Giovanna Nicora: Conceptualization, Methodology, Investigation, Writing-Review & Editing, Supervision; Enea Parimbelli: Writing-Review & Editing, Supervision; Tommaso Mario Buonocore: Writing-Review & Editing, Supervision; Roberto Bergamaschi: Validation, Writing-Review & Editing, Data Curation, Supervision; Eleonora Tavazzani: Writing-Review & Editing, Data Curation, Supervision; Arianna Dagliati: Writing-Review & Editing, Supervision, Funding acquisition; Riccardo Bellazzi: Writing-Review & Editing, Supervision, Funding acquisition;

## Acknowledgments

This work was supported by "Fit4MedRob- Fit for Medical Robotics" Grant B53C22006950001 This work was supported by the European Union's Horizon 2020 Brainteaser Project (GA101017598).